\documentclass[10pt,twocolumn,letterpaper]{article}

\usepackage{cvpr}              

\usepackage{graphicx}
\usepackage{amsmath}
\usepackage{amssymb}
\usepackage{booktabs}
\usepackage[table,xcdraw,dvipsnames]{xcolor}

\usepackage[pagebackref,breaklinks,colorlinks]{hyperref}

\usepackage[capitalize]{cleveref}
\crefname{section}{Sec.}{Secs.}
\Crefname{section}{Section}{Sections}
\Crefname{table}{Table}{Tables}
\crefname{table}{Tab.}{Tabs.}

\def\cvprPaperID{44} 
\def\confName{CVPR}
\def\confYear{2023}

\begin{document}

\title{Transfer Learning for Fine-grained Classification Using Semi-supervised Learning and Visual Transformers}

\author{Manuel Lagunas*\\
{\tt\small mlgns@amazon.com}
\and
Brayan Impata*\\
{\tt\small biimpata@amazon.com}
\and
Victor Martinez\\
{\tt\small vicmg@amazon.com}
\and
Virginia Fernandez\\
{\tt\small virfer@amazon.com}
\and
Christos Georgakis\\
{\tt\small georgak@amazon.com}
\and
Sofia Braun\\
{\tt\small brasofia@amazon.com}
\and
Felipe Bertrand\\
{\tt\small felipb@amazon.com}
}
\maketitle

\renewcommand{\thefootnote}{}

\begin{abstract}
Fine-grained classification is a challenging task that involves identifying subtle differences between objects within the same category. This task is particularly challenging in scenarios where data is scarce. Visual transformers (ViT) have recently emerged as a powerful tool for image classification, due to their ability to learn highly expressive representations of visual data using self-attention mechanisms. In this work, we explore Semi-ViT, a ViT model fine tuned using semi-supervised learning techniques, suitable for situations where we have lack of annotated data. This is particularly common in e-commerce, where images are readily available but labels are noisy, nonexistent, or expensive to obtain. Our results demonstrate that Semi-ViT outperforms traditional convolutional neural networks (CNN) and ViTs, even when fine-tuned with limited annotated data. These findings indicate that Semi-ViTs hold significant promise for applications that require precise and fine-grained classification of visual data.
\footnotetext{* Joint first authors.}
\end{abstract}


\section{Introduction}
\label{sec:intro}

In recent years, the development of deep neural networks has led to significant advancements in the field of computer vision~\cite{lecun2015deep}. One such architecture is the Visual Transformer (ViT)~\cite{dosovitskiy2020image}, which utilizes the self-attention mechanism to model long-range dependencies between image features. Unlike traditional convolutional neural networks (CNN)~\cite{he2016deep, simonyan2014very, goodfellow2016deep}, which rely on handcrafted hierarchical feature extraction, visual transformers can learn global spatial relationships among image features in a more efficient and effective manner. This has enabled them to outperform state-of-the-art methods on various visual recognition tasks~\cite{liu2021survey}. However, in real-world scenarios labeled data can be scarce and expensive to obtain. Therefore, semi-supervised learning (SSL)~\cite{zhu2005semi} has emerged as a powerful technique for leveraging unlabeled data to improve the performance of deep neural networks. CNN methods have significantly advanced the field~\cite{laine2016temporal, berthelot2019mixmatch, cai2021exponential, xie2020unsupervised, sohn2020fixmatch} while ViT architectures have only recently demonstrated promising results\cite{cai2022semi, xing2022svformer} with SSL.

In this paper, we investigate the effectiveness of SSL when used with ViT architectures. Specifically, we utilize the Semi-ViT architecture~\cite{cai2022semi} to conduct transfer learning for fine-grained classification of e-commerce data. The use of e-commerce data presents a unique advantage for SSL as unlabeled images are readily available. However, the labels associated are often noisy or absent altogether. Traditionally, this issue has been addressed through the use of manual curators, which can be costly and predominantly accessible for established marketplaces. In emerging markets, such as in Latin America, the scarcity of reliable labelled data poses an even greater challenge.

We collect three datasets from e-commerce data containing labeled and unlabeled images. We perform fine-grained classification on the neck style of a vest (\textit{Vest Neck Style)}, the pattern of a phone case (\textit{Phone Case Pattern)}, and the pattern of aprons and food bibs (\textit{Apron Food Bib Pattern)}. Each of the datasets contains 29K, 30K, and 33K labeled images, and 227K, 287K, 284K unlabeled images, respectively. Labels were gathered using crowd-sourced methods. 
We fine tune three different models, the well-known ResNet architecture~\cite{he2016deep}, a ViT, and a Semi-ViT architecture; all of them pretrained on ImageNet~\cite{deng2009imagenet}. For the ViT and Semi-ViT architectures, we additionally set different labeled data regimes where they are additionally fine-tuned using 25\%, 50\%, and 75\% of the labeled data for each of the datasets. In total, we train 9 different models for each task.
\section{Related Work}
\label{sec:related}

\paragraph{Visual Transformers}
Visual Transformers (ViT) have recently achieved state-of-the-art performance in many computer vision tasks~\cite{dosovitskiy2020image, liu2021swin, vaswani2021scaling}. They adapt self-attention mechanisms to the image domain, allowing to better capture long-range dependencies and contextual information. ViTs have been extended with knowledge distillation~\cite{touvron2021training}, using token-level and patch-level transformer layers~\cite{han2021transformer}, or progressively downsampling the image~\cite{yue2021vision}. A comprehensive review on ViTs can be found in the work of Khan et al.~\cite{khan2022transformers}. In this work, we explore and compare the performance of ViT architectures and traditional CNNs, fine-tuned with supervised and semi-supervised techniques for fine-grained classification.

\paragraph{Transfer Learning}
Transfer learning leverages pre-trained models and adapts them to new domains~\cite{zhuang2020comprehensive, oquab2014learning, lagunas2018transfer, sabatelli2018deep}. Yosinski et al.~\cite{yosinski2014transferable, yosinski2015understanding} investigate the transferability of features learned by deep neural networks on different tasks, demonstrating their effectiveness. Transfer learning has also been applied for object detection and semantic segmentation~\cite{girshick2014rich}, and in unsupervised domain adaptation~\cite{tzeng2017adversarial}. We use fine tuning methods in ResNet, ViT, and Semi-Vit~\cite{cai2022semi} architectures to perform fine-grained classification in three different datasets. 

\paragraph{Semi-Supervised Learning}
Semi-Supervised Learning (SSL) uses labeled and unlabeled data to improve model performance when labeled data is scarce~\cite{yang2022survey,zapata2020prediction,lagunas2019similarity}. It leverages intelligent data augmentation techniques paired with consistency regularization to improve performance~\cite{berthelot2019mixmatch, xie2020unsupervised,miyato2018spectral,zhang2017mixup}. Other approaches rely on pseudo-labelling~\cite{lee2013pseudo}, teacher-student models~\cite{qiao2018deep, tarvainen2017mean}, ensembles~\cite{laine2016temporal}, or adversarial training~\cite{kumar2017semi, miyato2018virtual}. We apply SSL to e-commerce images, where labeled data is expensive, but we can retrieve a large amount of unlabeled samples. We demonstrate the applicability of SSL and showcase its efficacy at reducing the need for labeled data.

\section{Data Collection}
\label{sec:data}

Our goal is to compare the performance of traditional CNNs and ViT architectures
on e-commerce images that allow us to have large amounts of unlabelled data. However, labelled data is noisy or nonexistent, therefore we would need crowd-sourcing tasks to label them.

\begin{figure}[t]
  \centering
   \includegraphics[width=\linewidth]{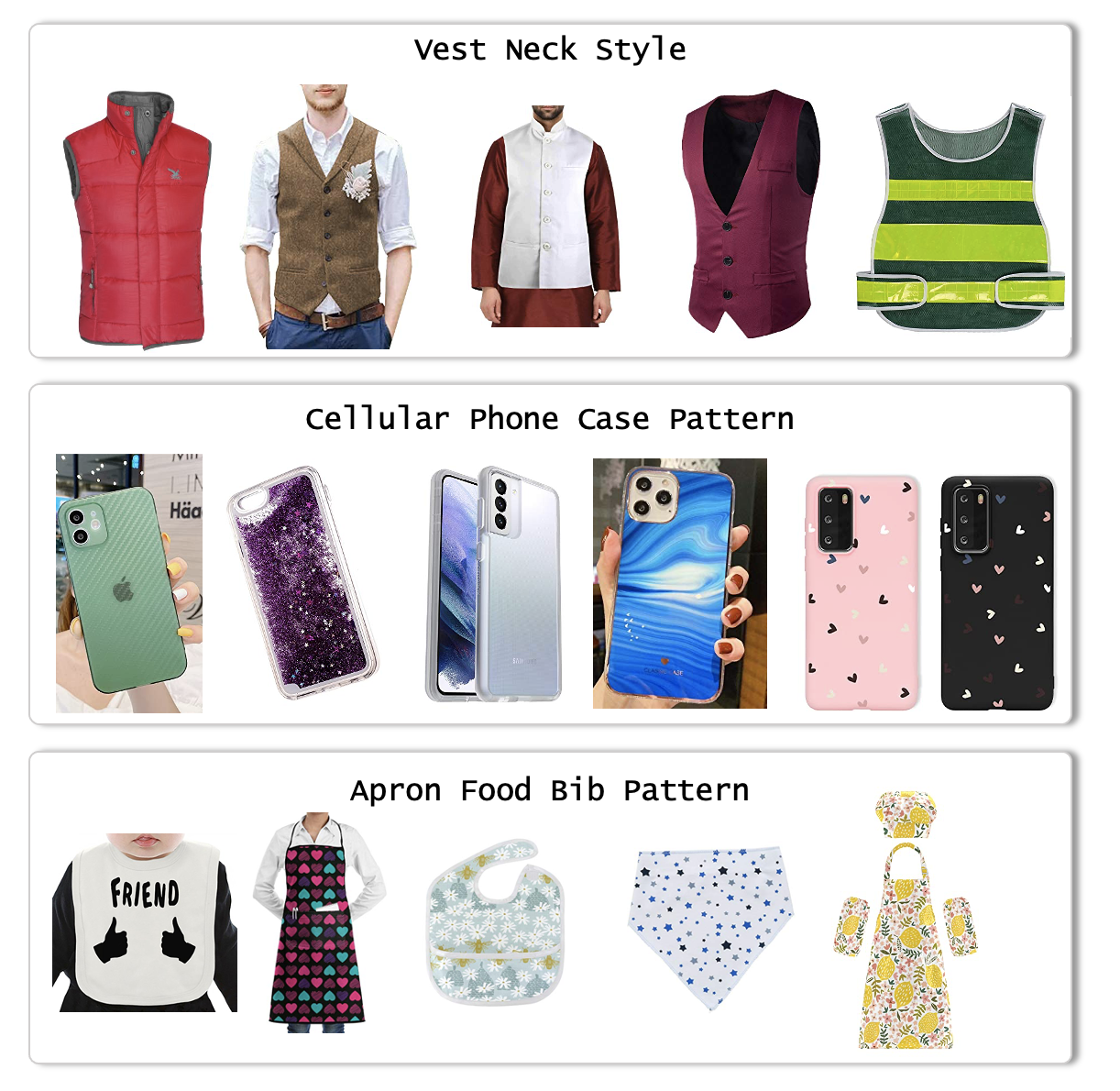}
   \caption{Examples of the images collected for our datasets. From top to bottom, we can see examples for \textit{Vest Neck Style}, \textit{Phone Case Pattern}, and \textit{Apron Food Bib Pattern}.}
   \label{fig:001_data_example}
\end{figure}

We use three different datasets to perform fine-grained classification: predict the neck style of vests (\textit{Vest Neck Style}), the pattern of phone cases (\textit{Phone Case Pattern}), and the pattern of aprons and food bibs (\textit{Apron Food Bib Pattern}). All images come from Amazon's marketplace. To label images we rely on Amazon Mechanical Turk. Each labelled image has answers from three different people. We consider an image as labelled if two or more labels are the same, otherwise, we would discard the labels and consider it unlabelled. To ensure the quality of the annotations, we only allowed workers that previously had passed a preliminary test task. Table~\ref{tab:dataset_summary} shows a summary on the dataset statistics. An example of images we collected for each of the three tasks can be seen in Figure~\ref{fig:001_data_example}.
\begin{table}[t]
    \centering
    \resizebox{\columnwidth}{!}{%
        \begin{tabular}{lccc}
         \rowcolor[HTML]{EFEFEF}  
        \multicolumn{4}{c}{\textsc{Dataset summary}} \\ 
        \toprule
        Dataset & Labelled & Unlabelled & Classes\\ 
        \midrule
        \textit{VestNeckStyle} & 29K & 227K & 13 \\ 
        \textit{PhoneCasePattern} & 37K & 287K & 21 \\ 
        \textit{ApronFoodBibPattern} & 39K & 284K & 26 \\ 
        \bottomrule
        \end{tabular}
    }
\caption{Summary of the number of images that are labelled, unlabelled, and the number of classes for each of the datasets that we collected from e-commerce data. All the datasets have a class named as \textit{other}. This class is used to label images that do not belong to the aforementioned data i.e., in \textit{Vest Neck Style}, an image showing something that is not a vest.}
  \label{tab:dataset_summary}
\end{table}

\section{Methodology}
\label{sec:method}

Our goal is to compare the performance of CNNs and modern ViT architectures. In addition, we will also study the influence of semi-supervised learning with ViT (Semi-ViT) during training using all available labeled data, and investigate the influence of more restrictive data regimes in ViT's and Semi-ViT's performance.

\paragraph{Models} We employ three different models: A CNN (a ResNet~\cite{he2016deep}) model, a ViT~\cite{dosovitskiy2020image} model, and a semi-ViT model~\cite{cai2022semi} (a ViT model trained also with SSL). 
For the CNN we use the well-know ResNet18~\cite{he2016deep} model. This model has shown remarkable performance across different computer vision tasks while being the less parameter-heavy ResNet model. In the ViT architecture, we have relied on Masked Autoencoders (MAE)~\cite{he2022masked} ViT-Base model. Following a similar approach to ResNet, this is the most parameter efficient MAE. For the Semi-ViT model~\cite{cai2022semi}, we also use the same MAE ViT-Base model. During the semi-supervised stage, an exponential moving average (EMA)-Teacher framework is adopted together with a probabilistic pseudo mixup~\cite{zhang2017mixup} method that allows for better regularization by interpolating unlabeled samples and their pseudo labels.

\paragraph{Data} The labeled data is separated into \textit{train}, \textit{validation}, and \textit{test}. Since we gather e-commerce data, the underlying label distribution is unknown, the \textit{train} and \textit{validation} sets do not have a uniform distribution of the labels across images. We sample the \textit{test} set to have the same number of images for all labels. The distribution of the data is roughly 75\%, 15\%, and 10\% for \textit{train}, \textit{validation}, and \textit{test} respectively.

\paragraph{Fine tuning} All three models have been pre-trained on ImageNet~\cite{deng2009imagenet}. 
We fine-tune every model on 100\% of the labeled samples. For ViT and Semi-ViT, we also experiment tuning them with 25\%, 50\%, and 75\% of the training data. The validation and test sets remains the same to have a fair comparison of performance. For Semi-ViT, we fine tune and later perform semi-supervised learning using both labeled and unlabelled data. We orchestrate the fine tuning of all models using \textit{SageMaker} \textit{g5} instances, we rely on their hyper-parameter tuner with Bayesian search~\cite{perrone2021amazon} configured to search hyper-parameters  around the original values given in the training code of the three models. On Average, training a model took 1h for ResNet, 4h for ViT, and 12h for Semi-ViT.

\section{Results}
\label{sec:results}

In this Section we present the results obtained by the fine-tuned models in all three tasks. Table~\ref{tab:global_results} shows results for all models in the \textit{Vest Neck Style}, \textit{Phone Case Pattern}, and \textit{Apron Food Bib Pattern} task. For each, we show the percentage of data used for training, the accuracy top-1 (Acc@1) and top-5 (Acc@5), and the cross entropy (CE) error. All values are obtained in the test set.


\begin{table}[t]
\centering
\begin{tabular}{lccc}
 \rowcolor[HTML]{EFEFEF}  
\multicolumn{4}{c}{\textsc{Vest Neck Style}} \\ 
\toprule
 Method & Acc@1 & Acc@5 & Loss (CE)  \\ 
 \bottomrule
 ResNet18/100\% & 79.736 & 98.775 & 0.615 \\ 
 \hline
 ViT/100\% & 81.244 & 99.246 & 0.631 \\ 
 ViT/75\% & 81.056 & 98.586 & 0.661 \\
 ViT/50\% & 80.584 & 98.586 & 0.704 \\ 
 ViT/25\% & 76.343 & 96.418 & 0.811 \\ 
 \midrule
 \rowcolor[HTML]{E8FFE8}  
 Semi-ViT/100\% & \textbf{85.297} & \textbf{99.246} & \textbf{0.549} \\
 Semi-ViT/75\%  & 83.129 & 98.963 & 0.573 \\ 
 Semi-ViT/50\%  & 81.433 & 98.115 & 0.637 \\ 
 Semi-ViT/25\%  & 81.244 & 97.455 & 0.679 \\ 
\bottomrule
\end{tabular}
\newline
\vspace{1.5em}
\newline
\begin{tabular}{lccc}
 \rowcolor[HTML]{EFEFEF}  
\multicolumn{4}{c}{\textsc{Phone Case Pattern}} \\ 
\toprule
Method & Acc@1 & Acc@5 & Loss (CE) \\ 
 \bottomrule
 ResNet18/100\% & 72.931 & 97.034 & 0.923\\ 
 \hline
 ViT/100\% & 79.005 & 98.613 & 0.685 \\
 ViT/75\%  & 77.810 & 98.135 & 0.732 \\
 ViT/50\%  & 77.571 & 97.513 & 0.758 \\
 ViT/25\%  & 74.175 & 97.131 & 0.871 \\

 \midrule
 \rowcolor[HTML]{E8FFE8}  
 Semi-ViT/100\% & \textbf{81.540} & \textbf{98.613} & \textbf{0.618} \\
 Semi-ViT/75\%  & 80.010 & 98.374 & 0.664 \\ 
 Semi-ViT/50\%  & 79.149 & 98.422 & 0.700 \\ 
 Semi-ViT/25\%  & 76.279 & 97.465 & 0.805 \\ 
\bottomrule
\end{tabular}
\newline
\vspace{1.5em}
\newline
\begin{tabular}{lccc}
 \rowcolor[HTML]{EFEFEF}  
\multicolumn{4}{c}{\textsc{Apron Food Bib Pattern}} \\ 
\toprule
Method & Acc@1 & Acc@5 & Loss (CE) \\ 
 \bottomrule
 ResNet18/100\%  & 73.766 & 95.598 & 0.987 \\ 
 \hline
 ViT/100\% & 78.079 & 97.688 & 0.739 \\
 ViT/75\%  & 78.301 & 97.821 & 0.757 \\  
 ViT/50\%  & 75.056 & 96.754 & 0.869 \\  
 ViT/25\%  & 69.898 & 94.309 & 1.070 \\ 
 \midrule
 \rowcolor[HTML]{E8FFE8}  
 Semi-ViT/100\% & \textbf{81.814} & 97.732 & \textbf{0.663} \\
 Semi-ViT/75\%  & 81.547 & \textbf{97.866} & 0.685 \\ 
 Semi-ViT/50\%  & 77.234 & 97.110 & 0.815 \\ 
 Semi-ViT/25\%  & 73.499 & 94.620 & 0.981 \\ 
 \bottomrule
\end{tabular}
\caption{Results using the ResNet18, ViT, and Semi-ViT models with different data regimes on the three datasets obtained from e-commerce sources (images) and crowd-sourced experiments (labels). We can see how Semi-ViT, the visual transformer model that also uses SSL techniques, obtains the best performance for all tasks (green highlighted). Moreover, we can see that the Semi-ViT model with scarce data regimes (e.g., 50\% of the training data), which is a common scenario in e-commerce, obtain performances comparable to ViT models trained with double the amount of data.}
  \label{tab:global_results}
\end{table}

We observe that the Semi-ViT model outperforms others in all three tasks, showcasing the benefits of SSL to achieve better generalization. On the other hand, ResNet18 is the model that obtains worst performances. Other interesting finding is that, in general, the performance of ViT/50\% is on par with Semi-ViT/25\%. Thus, using half the amount of annotated data. Similarly, the performance of Semi-ViT/50\%, is on par or superior to ViT/100\%.

\subsection{Performance per class}
If we analyze the performance per class for each of the three tasks using Semi-ViT/100\%, we observe that the models, in general, achieve high values of accuracy top-1, usually above 90\% per label.
We find performances below 50\% on the labels that are underrepresented in the training set. In the case of the \textit{Apron Food Bib Pattern} task, three worse performing labels: \textit{paisley}, \textit{patched}, and \textit{trellis}; show an accuracy top-1 of 18.18\%,  11.11\%, 40.00\% respectively. However, they only represent 0.17\%, 0.07\%, and 0.56\% of the training data. In the case of \textit{Phone Case Pattern}, we also observe \textit{paisley} as an underperforming class, nevertheless, in training it only represented 0.71\% of the data. One interesting case, is the \textit{inconclusive} label, representing phone cases that could not be labeled in any other class. The model shows a 45.78\% accuracy top-1 even when the class represents 3.17\% of the samples. We argue that this low performance comes due to the fact that this is a "wildcard" class in which we may find high degrees of variability between images. Last, for \textit{Vest Neck Style}, we only find the \textit{band collar} label to have low performance (17.65\%), this label only represented 0.82\% of the training data.

\subsection{Performance by marketplace} 
When downloading e-commerce data, we also stored  metadata such as the marketplace where this image was located. In Latin America, we obtained data from Mexico and Brazil. For \textit{Vest Neck Style}, we found the accuracy top-1 of the Semi-ViT/100\% model to be on par to other regions, with 85.18\% in Mexico, and 82.92\% in Brazil. For \textit{Phone Case Pattern}, the accuracy top-1 is 80.56\% in Mexico, and 60.00\% in Brazil. In this case, we found significantly less images in Brazil than in others. Therefore, we have few images to train, resulting in the model not properly learning the characteristics of the images for this marketplace. For \textit{Apron Food Bib Pattern} we find accuracy top-1 to be 78.88\% in Mexico, and 81.91\% in Brazil; on par with other marketplaces. 

\subsection{Influence on the amount of unlabeled data}
Since we use e-commerce data, obtaining new unlabeled images is a fairly easy process. Therefore, we investigate the effect of having additional unlabeled data in models' performance. We selected the Semi-ViT model using 50\% of the labeled data (Semi-ViT/50\%), and train it using SSL with 200\%, 300\%, and 400\% of the unlabeled data. In the end, we have additionally collected a total of 1139K unlabeled images for the \textit{Apron Food Bib Pattern} task.

We observe in Table~\ref{tab:moredata_results} that adding extra unlabeled images yields results that improve performance in the case of 200\% and 300\% of the available unlabeled images. However, it seems to collapse for 400\% where performance decreases and is on par with using 100\% of the unlabeled data. We argue that the labeled data represents a very small percentage compared to unlabeled data, thus, the latter data drives model's training yielding worse performance.
\begin{table}[t]
    \centering
    \begin{tabular}{cccc}
     \rowcolor[HTML]{EFEFEF}  
\multicolumn{4}{c}{\textsc{Additional unlabeled}} \\ 
    \toprule
    Unlabeled & Acc@1 & Acc@5 & Loss (CE) \\ 
    \midrule
    284K (100\%)  & 77.234 & \textbf{97.110} & 0.815 \\ 
    569K (200\%). & 78.301 & \textbf{97.110} & 0.786 \\ 
     \rowcolor[HTML]{E8FFE8} 
    854K (300\%)  & \textbf{79.146} & 96.843 & \textbf{0.781} \\ 
    1139K (400\%) & 77.679 & 96.843 & 0.801\\ 
    \bottomrule
    \end{tabular}
\caption{Results for \textit{Apron Food Bib Pattern} using the Semi-ViT/50\% model and increasing the amount of unlabeled data. We can see how the performance in terms of accuracy and loss improves while adding more data. However, it collapses with 400\% of unlabeled data. We hypothesize that this may be due to the percentage of labeled data being significantly smaller with respect to unlabeled data, thus, unlabeled data driving the training and not allowing to further generalize.}
  \label{tab:moredata_results}
\end{table}
\section{Conclusion}
\label{sec:conclusion}
We have run experiments with three models: ResNet, ViT, and Semi-Vit on three datasets that were collected from e-commerce data. The use of e-commerce data provided a realistic setting to assess the impact of SSL with a combination of labeled and unlabeled images. Our experiments showed that Semi-ViT can effectively leverage the benefits of SSL to improve the performance in fine-grained classification tasks compared to other architectures. We can see how Semi-ViT obtains accuracy values similar to ViT models that have trained with double the amount of data. Those results are shared across datasets showing promise to reduce the need for labeled data while maintaining high performance. Nevertheless, there are open venues worth exploring: One could aim to integrate the noise from the crowd-sourced labels into the loss function to create more robust models, or explore the creation of SSL techniques for multi-attribute models in which all three tasks would be performed by a single model.


{\small
\bibliographystyle{ieee_fullname}
\bibliography{egbib}
}

\end{document}